\documentclass[letterpaper]{article} 
\usepackage{aaai2026}  
\usepackage{times}  
\usepackage{helvet}  
\usepackage{courier}  
\usepackage[hyphens]{url}  
\usepackage{graphicx} 
\usepackage{natbib}  
\usepackage{caption} 
\usepackage{algorithm}
\usepackage{algpseudocode}
\usepackage{amssymb}
\usepackage{amsmath}

\usepackage{newfloat}
\usepackage{listings}
\usepackage{subcaption}
\usepackage{graphicx}
\usepackage{booktabs}
\usepackage{multirow} 
\DeclareCaptionStyle{ruled}{labelfont=normalfont,labelsep=colon,strut=off} 
\floatstyle{ruled}
\newfloat{listing}{tb}{lst}{}
\floatname{listing}{Listing}
\title{Escaping Optimization Stagnation: Taking Steps Beyond Task Arithmetic via Difference Vectors}
\author{
    Jinping Wang,
    Zhiqiang Gao\thanks{Corresponding author: zgao@wku.edu.cn.},
    Dinggen Zhang,
    Zhiwu Xie
}
\affiliations{
       College of Science, Mathematics and Technology, Wenzhou-Kean University\\
       \{1306325, zgao, 1336574, zxie\} @wku.edu.cn
}

\usepackage{bibentry}

\begin{document}

\maketitle

\begin{abstract}
Current methods for editing pre-trained models face significant challenges, primarily high computational costs and limited scalability. Task arithmetic has recently emerged as a promising solution, using simple arithmetic operations—addition and negation—based on task vectors which are the differences between fine-tuned and pre-trained model weights, to efficiently modify model behavior. However, the full potential of task arithmetic remains underexplored, primarily due to limited mechanisms for overcoming optimization stagnation. To address this challenge, we introduce the notion of difference vector, a generalized form of task vectors derived from the historical movements during optimization. Using difference vectors as directed perturbations, we propose the Difference Vector-based Anisotropic Scaling Iterative algorithm (DV-BASI) to enable a continuous optimization process for task arithmetic methods without relying on any additional modules or components. Notably, by leveraging escapability and directional advantages of difference vectors, the average performance on different tasks of the multi-task model merged by DV-BASI may even outperform models individually fine-tuned. Based on this observation, we extend the application of difference vectors to a feasible fine-tuning method for single-task models. On the practical side, DV-BASI allows expressive searching directions with few learnable parameters and forms a scalable framework.
We also integrate DV-BASI with task arithmetic methods and advanced optimization techniques to achieve state-of-the-art performance on both supervised and unsupervised evaluation protocols. Our code can be found at:\\
\url{https://github.com/smithgun2005/DVBASI-AAAI2026-Oral}
\end{abstract}


\section{Introduction}

Pre-trained models are essential in contemporary machine learning systems due to their efficiency and transferability. Editing models after pre-training is widely recognized as an effective way to enhance model performance on specific downstream tasks \cite{DBLP:conf/cvpr/WortsmanIKLKRLH22, DBLP:journals/pieee/ZhuangQDXZZXH21, DBLP:conf/nips/MatenaR22}, mitigate undesired behaviors \cite{DBLP:conf/nips/SanturkarTEBTM21, DBLP:conf/acl/RibeiroL22, DBLP:conf/emnlp/MurtyMLR22}, align models with human preferences \cite{DBLP:journals/corr/abs-2112-00861, DBLP:conf/nips/Ouyang0JAWMZASR22,DBLP:journals/corr/abs-2209-00731}, or incorporate new information \cite{DBLP:conf/emnlp/CaoAT21, DBLP:conf/iclr/MitchellLBFM22, DBLP:conf/icml/MitchellLBMF22}.
However, traditional editing approaches, which rely on expensive joint fine-tuning across multiple tasks \cite{DBLP:conf/acl/VuLCAC22} and human feedback \cite{matthews1975comparison}, face limitations in scalability and accessibility. Moreover, optimizing models for downstream tasks often comes at the expense of diminished pre-training performance or zero-shot accuracy \cite{DBLP:conf/nips/GaripovIPVW18, DBLP:conf/iclr/LoshchilovH19, DBLP:conf/ijcnn/StallkampSSI11}.

Recently, innovative research on task arithmetic has introduced cost-effective and scalable model editing techniques \cite{task_arithmetic, TIES-Merging, AdaMerging, Tangent_Space, tau-jp, aTLAS}. By leveraging the concept of task vector that is defined as the element-wise difference between the weights of fine-tuned and pre-trained models, task arithmetic can modify various models through simple arithmetic operations on these vectors  \cite{task_arithmetic}. Specifically, negating a task vector can eliminate undesirable behaviors on specific tasks (task negation), while adding task vectors from different tasks can lead to the creation of a multi-task model that performs well on multiple tasks simultaneously (task addition).
\begin{figure*}[t]
    \centering    \includegraphics[width=\linewidth]{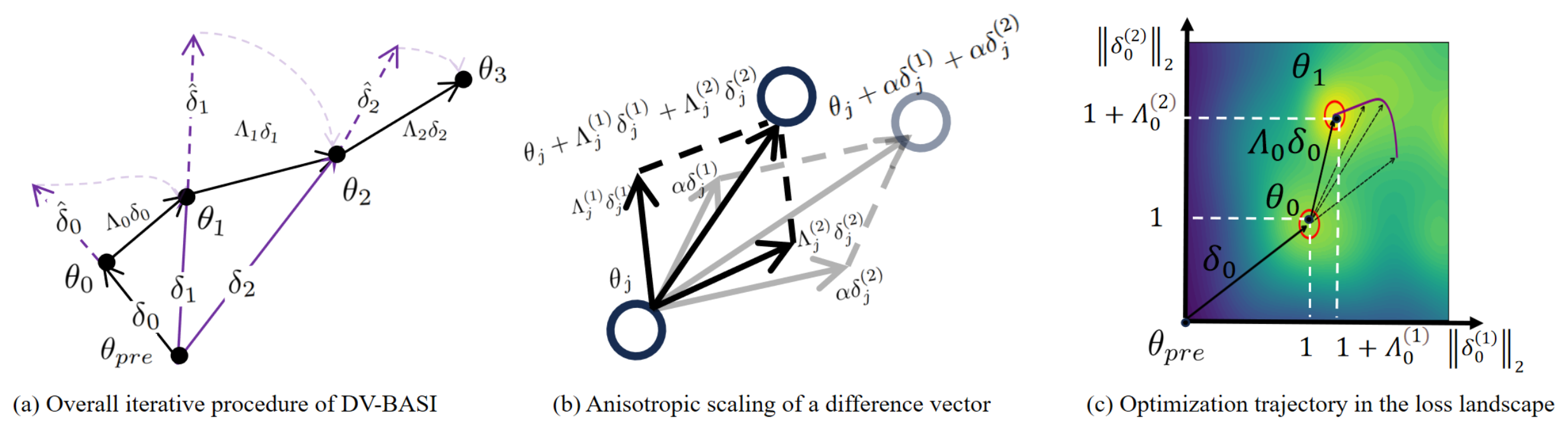}
    \caption{
The overall iterative procedure of DV-BASI is illustrated in (a). Starting from pre-trained weights \(\theta_{\text{pre}}\), the model initially reaches a local optimum \(\theta_0\) during the first major optimization step. At each local optimum \(\theta_j\), DV-BASI computes difference vectors \(\delta_j\) (indicated by purple arrows), which provide directional guidance for further optimization (\(\hat{\delta}_j\) denotes the directional vector of \(\delta_j\)). Based on these difference vectors, we apply anisotropic scaling matrices \(\Lambda_j\) to create more flexible exploration directions, aiming to find a potentially better global solution.
(b) provides a detailed illustration of the anisotropic scaling process for a difference vector. Assume each difference vector has two parameter blocks \(\delta_j = (\delta_j^{(1)}, \delta_j^{(2)})\). Each block is independently scaled by the anisotropic matrix \(\Lambda_j\) (where \(\Lambda_j = (\Lambda_j^{(1)}, \Lambda_j^{(2)})\)), which offers more expressive searching directions compared to using a scalar scaling coefficient \(\alpha\) \cite{aTLAS}.
(c) visualizes the iterative optimization path of DV-BASI in a loss landscape. It demonstrates how difference vectors function as directed perturbations, effectively helping model weights escape from the current local optima (red circles) to continue searching anisotropically (the purple line represents the anisotropic scaling trajectory of DV-BASI, based on gradient descent) for a potentially better solution in the parameter space.}
  \label{fig:three_images}
\end{figure*}
Recent advances on linearized task vectors deepen the theoretical understanding on task arithmetic by addressing the interference among task vectors. Through techniques based on model linearization via the neural tangent kernel approximation \cite{Tangent_Space} and $\tau$-Jacobian product regularization ($\tau$Jp Reg) \cite{tau-jp} during the \textit{model pre-training} stage, the linearized task vectors can be produced with less weight disentanglement error.

Although recent studies have advanced our understanding of task arithmetic, current approaches for designing task vector combination strategies have not yet realized the full potential of task arithmetic. Ideally, a merged multi-task model edited through task arithmetic is expected to achieve performance comparable to that of individually fine-tuned single-task models. However, due to the limited expressive power of combination coefficients learned via coarse grid search \cite{task_arithmetic, TIES-Merging}, this goal remains elusive in practice. Although current finer-grained \textit{Parameter-Efficient Fine-Tuning (PEFT)} task vector combination methods based on block-wise optimization \cite{aTLAS, AdaMerging} are addressing this issue, they are still fundamentally constrained by their single-step nature. Specifically, optimization often stops prematurely when model parameters become trapped in local optima where gradients vanish, thus impeding further exploration. Therefore, analogous to traditional parameter optimization methods, developing a multi-step optimization approach that can efficiently escape local optima and realize continuous optimization to find a better global solution is a crucial task.

To address this challenge, we propose a difference vector-based anisotropic scaling iterative algorithm (DV-BASI) to achieve continuous exploration in the parameter space, as illustrated in Figure~\ref{fig:three_images}. 
We extend the concept of task vectors to a more general difference vector, defined as the element-wise difference between the weights of a model in any arbitrary state during training and those of the pre-trained model. 
Similar to task vectors as knowledge carriers, the difference vector, as a cumulative result of previous optimizations, contains information of historical movements about the model weights from the training process.
Through theoretical and empirical analysis, we demonstrate that difference vectors enable continuous model optimization with the following merits:
$\text {(i)}$ \textit{Escapability and Directional Advantage}: When model weights are trapped in a local optimum, the updated difference vector at that point acts as a directed perturbation, effectively helping the model weights escape the current critical point and continue searching for a potentially better solution.
$\text {(ii)}$ \textit{Component-Free Continuity}: Continuous exploration in the parameter space relies solely on the updates of the difference vector, without depending on additional components such as adapters \cite{DBLP:conf/icml/HoulsbyGJMLGAG19}, prompts \cite{jia2022visual}, or LoRA \cite{DBLP:conf/iclr/HuSWALWWC22}.

We demonstrate that DV-BASI is a scalable multi-step task arithmetic framework. Adhering to the standard evaluation protocols of task arithmetic \cite{task_arithmetic,Tangent_Space,tau-jp} and its extended PEFT paradigm \cite{aTLAS,AdaMerging}, our framework can seamlessly integrate with existing task arithmetic techniques (for example $\tau$JP and aTLAS) and adapt both unsupervised and supervised learning settings, delivering state-of-the-art (SOTA) performance. 

To further highlight its scalability from an optimization standpoint, we propose a Multi-Objective Optimization (MOO) strategy that treats each task as an independent objective as a case study to extend our learning framework.

The contributions of methods are as follows:
(1) We extend the learning paradigm of task arithmetic to a multi-step approach called DV-BASI  by employing difference vectors. DV-BASI can effectively enhance task arithmetic performance and realize state-of-the-art (SOTA) performance under both supervised and unsupervised settings.  
Empirical analyses and theoretical explanations demonstrate that difference vectors possess escapability and directional advantages, enabling conventional methods to escape local optima for further improvement. Additionally, the component-free nature of difference vectors promotes continuous exploration in the parameter space.
(2) Leveraging the benefits of our difference vectors, we expand the application scope of task arithmetic to further enhance the performance of already fine-tuned single-task models.
(3) The proposed DV-BASI is a novel and scalable framework that can easily integrate with conventional task arithmetic methods. This integration effectively unleashes their potential, resulting in better performances. 

\section{Models and Difference Vectors}
Investigations into task arithmetic, as initially explored in \cite{task_arithmetic}, have revealed intriguing attributes of task vectors across diverse models. Following the established setting of aTLAS \cite{aTLAS}, this study focuses on the CLIP \cite{CLIP} model, leveraging its extensive availability and manageable scale to facilitate a deeper analysis. Specifically, we derive task vectors through fine-tuning the image encoder while preserving the text representations. This method ensures that image encoders fine-tuned on distinct datasets produce features within a unified representational space, thanks to a common text encoder. As a result, task vectors from these fine-tuned encoders can be more seamlessly integrated to create a cohesive multi-task model.

Formally, let the CLIP image encoder be represented as \( f: \mathcal{X} \times \Theta \rightarrow \mathcal{Z} \), where for an input image \( x \in \mathcal{X} \) and parameters \( \theta \in \Theta \), \( z = f(x; \theta) \) denotes the learned latent representation of the input image. Let the weights of a pre-trained model be \( \theta_{pre} \), and the weights of its fine-tuned version be \( \theta^{(i)}_{ft} \), with \( i \in \mathbb{N}^+ \), where \( i \) indexes a dataset \( \mathcal{D}^{(i)} \). Following Ilharco et al. \cite{task_arithmetic}, we define a task vector as \( \tau^{(i)} = \theta^{(i)}_{ft} - \theta_{pre} \).

Generally, task arithmetic methods merge task vectors into a multi-task model by developing efficient weighting schemes to combine them. Given the high-dimensional and complex nature of model parameter spaces, merged models often encounter local optima, where standard training fails to significantly reduce loss or enhance model performance. Garipov et al. \cite{DBLP:conf/nips/GaripovIPVW18} have shown that despite the complexity of the loss surfaces of deep neural networks, optimal points are not isolated and can be connected through a simple low-loss path. This finding demonstrates the rationality to continue searching for other possible better solutions as long as along the proper direction.

The difference vector is instrumental in constructing a multi-step approach to assist task arithmetic in escaping local optima for further enhancement. When a merged model becomes temporarily trapped in a local optimum \(\theta^{*}\), the difference vector \(\delta\) can be defined as the element-wise difference between \(\theta^{*}\) and \(\theta_{pre}\), that is \(\delta = \theta^{*} - \theta_{pre}\). Since task vectors act as carriers of knowledge \cite{aTLAS}, difference vectors, as generalized task vectors, also encapsulate the historical model knowledge from the pre-trained to the local optima.

\section{Escaping Optimization Stagnation via DV-BASI}


\begin{algorithm}[t]
\caption{\textbf{DV-BASI}}
\textbf{Input:} Pre-trained weights $\theta_{\text{pre}}$, initial weights $\theta_{0}$ (obtained by pre-defined model merging methods), learning rate $\eta$, number of iterations $M$ \\
\textbf{Output:} Final weights $\theta_{M}$
\begin{algorithmic}[1]
    \For{$m = 1$ to $M$}
        \State $\delta_{m-1} \gets \theta_{m-1} - \theta_{\text{pre}}$\quad// Difference vector
        \State Initialize $\Lambda_{m-1}^{(0)}$
        \For{$t = 0,1,\dots$ until early stopping}
            \State $\Lambda_{m-1}^{(t+1)} \gets \Lambda_{m-1}^{(t)} - \eta \nabla_{\Lambda_{m-1}^{(t)}} \mathcal{L}(\theta_{m-1}^{(t)})$\quad// Solve Eq.(5)
        \EndFor
        \State Let $\Lambda_{m-1}^*$ be the converged scaling matrix
        \State $\theta_{m} \gets \theta_{m-1} + \Lambda_{m-1}^* \delta_{m-1}$
    \EndFor
\end{algorithmic}
\label{algorithm}
\end{algorithm}




DV-BASI is an extensible multi-step task arithmetic framework that can be used to optimize the models merged in the previous iteration.
%
Starting from the initial optimal point \( \theta_0 \) the corresponding initial difference vector \( \delta_0 \) is defined as the element-wise difference between \( \theta_0 \) and \( \theta_{\text{pre}} \). 
\textit{Notably, the initial optimal point \( \theta_0 \) represents the parameters of a model that have been merged by pre-defined methods, such as aTLAS and $\tau$JP reg as shown in our Experiments.
}
Here, the first iteration is formulated as:
\begin{equation}
    \delta_0 = \theta_0 - \theta_{\text{pre}}, \quad \theta_1 = \theta_0 + \Lambda_0 \delta_0,
\end{equation}
where \( \theta_1 \) represents the updated model parameters, and \( \Lambda_0 \) is a learnable anisotropic scaling matrix, which will be detailed in Equations~\ref{eq:coefficient1} and \ref{eq:coefficient2}.

During each iteration process, if the model's weights get stuck in a local optimal solution that is difficult to break through (manifested as no improvement in accuracy over several epochs), the best-performing parameters will be selected as the starting point for the subsequent iteration.
Here, the difference vector is updated to serve as a directed perturbation, propelling the model parameters away from this local optimum to facilitate ongoing exploration. Generally, in the \( (j+1) \)-th iteration, the difference vector \( \delta_j \) is updated based on \( \theta_j \), and the update for the next optimum \( \theta_{j+1} \) is given by:
\begin{equation} \label{ODE}
    \delta_j = \theta_j - \theta_{\text{pre}}, \quad \theta_{j+1} = \theta_j + \Lambda_j \delta_j.
\end{equation}

\begin{figure*}[!t]
  \centering
  \begin{subfigure}[b]{0.305\textwidth}
    \centering
    \includegraphics[width=\linewidth]{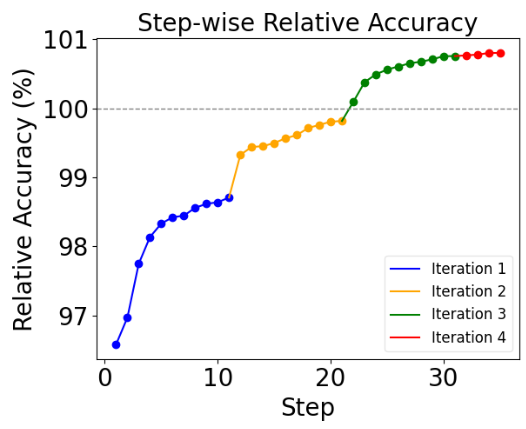}
    \caption{}
    \label{fig:sub-b}
  \end{subfigure}\hfill
    \begin{subfigure}[b]{0.295\textwidth}
    \centering
    \includegraphics[width=\linewidth]{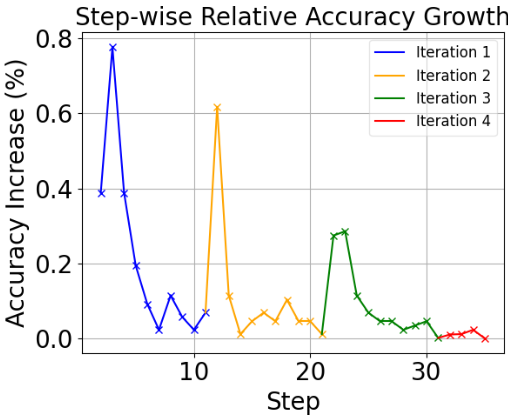}
    \caption{}
    \label{fig:sub-b}
  \end{subfigure}\hfill
  \begin{subfigure}[b]{0.4\textwidth}
    \centering
    \includegraphics[width=\linewidth]{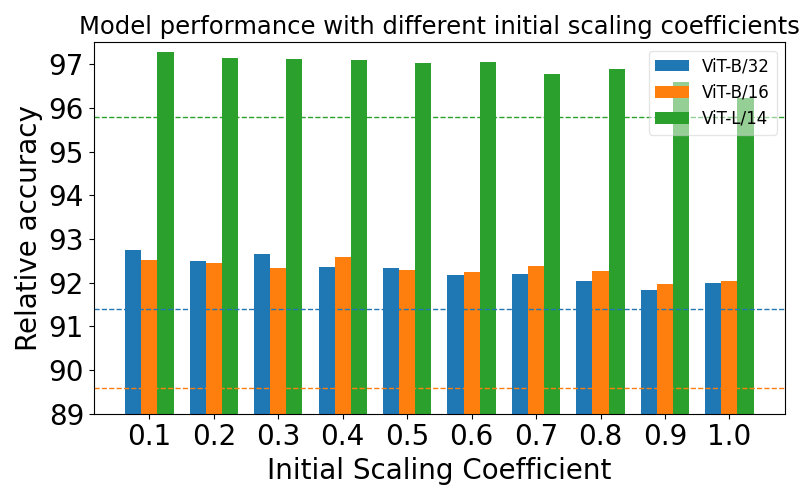}
    \caption{}
    \label{fig:sub-c}
  \end{subfigure}
  \caption{Figure (a) and (b) show the stepwise relative accuracy of supervised model merging (using ViT-B/32 as pre-trained backbone) and its growth within 4 DV-BASI iterations. Figure (c) compares the unsupervised model merging performance of 10 different initial scaling coefficients (0.1 to 1.0) among 3 pre-trained backbones (ViT-B/32, ViT-B/16, and ViT-L/14).}
  \label{fig:three-subfigs}
\end{figure*}

\begin{table*}[t]
  \footnotesize
  \caption{Unsupervised model merging performance comparison between different types of perturbations and scaling strategies. \(R_{j}\) denote the unit random vector.}
  \label{tab:vit_comparison}
  \centering
  \footnotesize
  \resizebox{0.55\textwidth}{!}{ 
    \begin{tabular}{lccc}
      \toprule
      Method      & ViT-B/32 & ViT-B/16 & ViT-L/14 \\
      \midrule
      Initial weights ($\theta_0$)    & 82.9     & 82.8     & 90.2     \\
      \midrule
      Random perturbation ($\theta_{j+1}=\theta_j+\Lambda_j\| \delta_j\|R_j$)     & 5.6     & 6.3     & 4.2     \\
      Isotropic (scalar) scaling ($\theta_{j+1}=\theta_j+\alpha_j\delta_j$)         & 83.3     & 83.6     & 90.3    \\
      Anisotropic scaling ($\theta_{j+1}=\theta_j+\Lambda_j\delta_j$) & \textbf{83.9}     & \textbf{85.3}     & \textbf{90.8}     \\
      \bottomrule
    \end{tabular}
  }
\end{table*}

To effectively explore the next optimum in each iteration, an anisotropic scaling mechanism is applied to the difference vectors, enabling flexible and controllable exploration within the parameter space. Typically, parameters across different layers of neural networks have distinct roles and functionalities. Inspired by Zhang et al. \cite{aTLAS}, instead of using a scalar \( \alpha \) to scale the difference vector, we decompose the difference vector into parameter blocks \( \delta = (\delta^{(1)}, \delta^{(2)}, ..., \delta^{(n)}) \) and assign each block an independent learnable scaling coefficient for anisotropic exploration in the parameter space. Consequently, we introduce a block diagonal scaling matrix \( \Lambda \):
\begin{equation}
\Lambda = \begin{bmatrix}
\lambda^{(1)} I^{(1)} & \cdots & 0 \\
\vdots & \ddots & \vdots \\
0 & \cdots & \lambda^{(n)} I^{(n)}
\end{bmatrix},
    \label{eq:coefficient1}
\end{equation}
where \( \lambda^{(i)} \) denotes the coefficient for each block, and \( I^{(i)} \) represents the identity matrix. This results in anisotropic scaling of the difference vector, expressed as:
\begin{equation}
    \Lambda \delta = (\Lambda^{(1)} \delta^{(1)}, \Lambda^{(2)} \delta^{(2)}, ..., \Lambda^{(n)} \delta^{(n)}).
    \label{eq:coefficient2}
\end{equation}

The anisotropic scaling matrix \( \Lambda \) is the sole learnable parameter within each iteration. Assuming a supervised learning context, the optimization problem for the \( (j+1) \)-th iteration is formulated as:
\begin{equation} \label{object}
\arg\min_{\Lambda_j}\;
\mathbb{E}_{(x,y)\sim \mathcal{D} }
\!\Bigl[
\mathcal{L}\!\bigl(f(x;\,\theta_j+\Lambda_j\delta_j),\,y\bigr)
\Bigr],
\end{equation}
where \( \mathcal{L} \) denotes the cross entropy loss function for the target task, and \( \mathcal{D} = \{ \mathcal{D}^{(i)} \}_{i=1}^{n} \) represents \( n \) target datasets from which \( (x, y) \) is drawn. 
Notably, when adapting DV-BASI to an unsupervised learning scenario, the detailed unsupervised loss function can be found in the Appendix.
The iterative learning process is introduced in Algorithm~\ref{algorithm}.

The DV-BASI is a scalable framework, that allows an advanced optimization paradigm, can be flexibly integrated into other task arithmetic approaches.
We introduce a Multi-Objective Optimization (MOO) approach, treating each task as a distinct objective and facilitating balanced optimization across multiple tasks. By employing the concept of Pareto optimality, 
we utilize the Multiple Gradient Descent Algorithm (MGDA) \cite{pareto} to ascertain the most balanced optimization direction, effectively minimizing losses across tasks as parallel objectives. The specifics of our MOO algorithm are elaborated in the Appendix.

\section{Empirical Analysis}

Difference vector is the cumulative result of all successful optimization steps taken so far, containing information about directions that have successfully reduced the loss in the previous training process. By applying difference vectors as perturbations, we can deliberately push the weights along the direction known to be effective, thereby escaping the current equilibrium in a guided manner. In this section, we will further discuss the escapability of the difference vectors and directional advantages compare to random perturbations.

\begin{table*}[t]
  \centering
    \footnotesize
  \caption{Performances of task negation averaged across eight datasets. 
  All our results maintain at least 95\% of the pre-trained accuracy on the control dataset.
  Results marked with an asterisk * indicate the supervised setting.}
  \label{tab:negation-performance}
  \resizebox{0.75\textwidth}{!}{  
    \begin{tabular}{llcccccc}
      \toprule
      \multirow{2}{*}{T.V.} & \multirow{2}{*}{Methods} & \multicolumn{2}{c}{ViT-B/32} & \multicolumn{2}{c}{ViT-B/16} & \multicolumn{2}{c}{ViT-L/14} \\
      \cmidrule(lr){3-4} \cmidrule(lr){5-6} \cmidrule(lr){7-8}
       & & Target ($\downarrow$) & Control ($\uparrow$) & Target ($\downarrow$) & Control ($\uparrow$) & Target ($\downarrow$) & Control ($\uparrow$) \\
      \midrule
    
      n/a & Pre-trained      & 48.1 & 63.4 & 55.5 & 68.3 & 64.9 & 75.5 \\
      
      \midrule
       \textbf{Unsupervised}\\
      \midrule
      \midrule
 
      \multirow{4}{*}{Std.}
      & Task arithmetic    & 24.0 & 60.9 & 21.3 & 65.4 & 19.0 & 72.9 \\
      & Ties-Merging       & 21.8 & \textbf{61.7} & 24.3 & \textbf{67.0} & 26.6 & \textbf{74.4} \\
      & aTLAS       &23.3 & 60.7 & 21.0 & 65.0 & 17.8 & 73.2 \\
       &\textbf{ aTLAS  + DV-BASI }& \textbf{20.3$^{-1.5}$} & 60.2 & \textbf{20.2$^{-0.8}$} & 65.0 & \textbf{15.2$^{-2.6}$} & 72.5 \\       
        \midrule
     \multirow{3}{*}{Lin.}
      & Task arithmetic    & 10.9 & 60.8 & 11.3 & 64.8 & 7.9 & 72.5 \\
      & $\tau$JP reg       & 6.7 & 60.8 & 4.7 & \textbf{66.0} & 3.7 & 73.0 \\
       & \textbf{$\tau$JP reg + DV-BASI} & \textbf{5.7$^{-1.0}$} & \textbf{60.8} & \textbf{4.4$^{-0.3}$} & 65.0 & \textbf{3.6$^{-0.1}$} & \textbf{73.2} \\  
       \midrule
       
       \textbf{Supervised}\\
        \midrule
        \midrule
        \multirow{2}{*}{Std.}
       & aTLAS$^*$       & 19.4 & \textbf{61.2} & 18.1 &\textbf{65.8} & 17.8 & \textbf{73.3} \\
       & \textbf{aTLAS$^*$ + DV-BASI$^*$}       & \textbf{10.7$^{-8.7}$} & 60.6 & \textbf{14.5$^{-3.6}$} & 64.9 & \textbf{12.6$^{-5.2}$} & 72.5 \\
      \midrule
      \multirow{3}{*}{Lin.}
 
        & aTLAS$^*$              & 11.1 & {61.0} & 10.2 & \textbf{65.6} & 12.6 & 73.1 \\
         &\textbf{aTLAS$^*$+ DV-BASI$^*$}  & {9.5$^{-1.6}$} & \textbf{61.3} & {8.4$^{-1.8}$} & 65.0 & {11.2$^{-1.4}$} & 73.2\\

       &\textbf{$\tau$JP reg + DV-BASI$^*$ } & \textbf{4.1$^{-7.0}$} & {61.0} & \textbf{3.6$^{-6.6}$} & 65.4 & \textbf{2.1$^{-10.5}$} & \textbf{73.6} \\
      \bottomrule
    \end{tabular}
  }
  \label{tab:negation}
\end{table*}

\paragraph{Escapability: Difference vectors, functioning as directed perturbations, can effectively push the model weights away to escape the optimization stagnation points.} 
To empirically verify its escapability, we illustrate the stepwise relative accuracy (accuracy of merged model divided by that of the fine-tuned models) and its growth interval, in Figures~\ref{fig:three-subfigs} (a) and (b), where all results are produced under a supervised learning setting. We experiment with 4 iterations, within the same iteration, the accuracies are marked with the same color, and the value of each point represents the result obtained after each training epoch. 
In (a), after applying DV-BASI, the relative accuracy iteratively increases and finally outperforms 100\%. This observation means that the performances of the merged model exceed those of fine-tuned models.
In (b), we can observe that during the early stages of each iteration, when the difference vectors are updated, the performances improve significantly, but as training progresses, the improvement gradually stagnates. When we update the difference vector at the beginning of each iteration, the previously stagnated optimization can resume. These results explicitly illustrate that our method effectively helps the model to escape from local stagnation and results in continuous improvement.

Unlike random perturbations, difference vectors as knowledge carriers contain knowledge learned from previous training steps. 

Denote $k$ as the step index, $K$ as the total number of training steps to reach the local optimal point $\theta^*$, and $\theta^{(k)}$ represents the model weights at $k_{th}$ step, we have:
\begin{equation}
\delta=\theta^*-\; \theta_{\mathrm{pre}}
\;=\;\sum_{k=1}^{K}\Delta\theta^{(k)}, \quad 
 \Delta\theta^{(k)} = \theta^{(k)} - \theta^{(k-1)}.
\end{equation}

To further investigate the effect of the initial scaling coefficient \(\alpha\) for each parameter block on the performance, we evaluate the merged model's under an unsupervised setting with different initial \(\alpha\) values (from 0.1 to 1.0).
As shown in Figure~\ref{fig:three-subfigs} (c), we observe that, across the range of initial scaling coefficients, DV-BASI consistently improves model performance compared to the initial pre-trained weights. 
This suggests that as long as the perturbation is applied in a proper direction, DV-BASI is insensitive to \( \alpha \) and it is possible to converge to a better solution. 
Intuitively, the above observation may mainly be attributed to the directional advantage of the difference vector, which leads to an exploration of the following directional properties of difference vectors.

\paragraph{Directional Advantage: Compared to random perturbations, using difference vectors as perturbations has a directional advantage. They always point in the direction where model weights improved in previous optimization steps.}
To demonstrate the advantage of perturbing along the direction of difference vectors, we compare the model performance with that of a model using random perturbations (with the same magnitude as our difference vectors) under an unsupervised model merging setting. Meanwhile, we also compare model performances under anisotropic and isotropic scaling strategies, respectively.  
As observed in Table~\ref{tab:vit_comparison}, unlike difference vectors, random perturbations lead to a catastrophic drop in performance, indicating the necessity of perturbing along the direction of difference vectors. \textit{A more theoretical perspective on why this phenomenon exists (Random perturbations can degrade model performance, whereas difference vectors contribute to continued optimization) is illustrated in the Appendix.} Meanwhile, the results show that compared to isotropic scaling, scaling the difference vector anisotropically is more likely to realize a better model performance due to its flexibility in searching directions.

Our difference vector functions similarly to the momentum term in traditional gradient-based optimization, where the update direction from the historical optimization steps serves as an inertial force to help the model escape from the current local minimum (where gradients vanish). However, like momentum, the direction provided by the difference vector does not guarantee convergence to a better global optimum. 
In fact, in the context of task arithmetic, referencing a global historical update direction is often significantly more effective than using a random one \ref{tab:vit_comparison} with significantly higher probability of reaching a better solution.

\paragraph{Computational Efficiency}

\begin{table*}[t]
  \centering
  \footnotesize
  \caption{Performances of task addition averaged across eight datasets. We report absolute accuracy (Abs.) and relative accuracy (Rel.) with respect to the average accuracy of model fine-tuned on single tasks. Results marked with an asterisk * indicate the supervised setting.}
  \label{tab:combined-addition-performance}
  \resizebox{0.75\textwidth}{!}{%
    \begin{tabular}{llcccccc}
      \toprule
      T.V. & Methods & \multicolumn{2}{c}{ViT-B/32} & \multicolumn{2}{c}{ViT-B/16} & \multicolumn{2}{c}{ViT-L/14} \\
      \cmidrule(lr){3-4}\cmidrule(lr){5-6}\cmidrule(lr){7-8}
      & & Abs. ($\uparrow$) & Rel. ($\uparrow$) & Abs. ($\uparrow$) & Rel. ($\uparrow$) & Abs. ($\uparrow$) & Rel. ($\uparrow$) \\
      \midrule
    
      n/a 
      & Pre-trained           & 48.1 &     –  & 55.5 &     –  & 64.9 &     –   \\
      \midrule
      \textbf{unsupervised}\\
      \midrule
      \midrule
      \multirow{5}{*}{Std.}
      & Task arithmetic       & 70.1 & 77.2 & 73.6 & 79.9 & 82.9 & 87.9  \\
      & Ties-Merging          & 74.2 & 84.8 & 78.6 & 87.6 & 85.0 & 91.9  \\
      & AdaMerging            & 80.1 & 88.5 & 82.9 & 89.7 & 90.8 & 96.4  \\
      & aTLAS         & 82.9 & 91.4 & 82.8 & 89.6 & 90.2 &  95.8 \\
      & \textbf{aTLAS + DV-BASI}           & \textbf{83.9$^{+1.0}$} & \textbf{92.8$^{+1.4}$} & \textbf{85.3$^{+2.5}$} & \textbf{92.5$^{+2.9}$} & \textbf{90.8$^{+0.6}$} &\textbf{96.4$^{+0.6}$} \\
         \midrule
      \multirow{3}{*}{Lin.}
      & Task arithmetic       & 74.7 & 85.2 & 77.5 & 86.2 & 84.8 & 91.9  \\
  
      & $\tau$JP reg        & 84.5 & 97.6 & 87.6 & 98.1 & 90.8 & 99.0  \\
       & \textbf{$\tau$JP reg + DV-BASI}     & \textbf{86.7$^{+2.2}$} & \textbf{99.8$^{+2.2}$}  & \textbf{88.0$^{+0.4}$} & \textbf{98.6$^{+0.5}$} & \textbf{91.8$^{+1.0}$} & \textbf{100.1$^{+1.1}$} \\
       \midrule
       \textbf{Supervised}\\
      \midrule 
      \midrule
      \multirow{3}{*}{Std.}
    & aTLAS$^*$                 & 84.1 & 92.8 & 82.9 & 89.7 & 91.4 & 97.1  \\
      & \textbf{aTLAS$^*$  + DV-BASI$^*$}             & {85.9$^{+1.8}$} & {94.8$^{+2.0}$} &{87.2$^{+4.3}$} & {94.6$^{+4.9}$} & {91.6$^{+0.2}$} & {97.4$^{+0.5}$} \\
      & \textbf{aTLAS$^*$  + DV-BASI$^*$  + MOO$^*$}         & \textbf{86.2$^{+2.1}$} & \textbf{95.1$^{+2.3}$} & \textbf{87.7$^{+4.8}$} & \textbf{95.1$^{+5.4}$} & \textbf{91.8$^{+0.4}$} & \textbf{97.6$^{+0.5}$} \\ 

     \midrule
      \multirow{3}{*}{Lin.} 
          & aTLAS$^*$                 & 83.4 & 95.4 & 85.4 & 95.1 & 88.7 & 96.1  \\
            & \textbf{aTLAS$^*$ + DV-BASI$^*$}                 & 86.6$^{+3.2}$ & 99.1$^{+3.7}$ & 87.5$^{+2.1}$ & 97.4$^{+2.3}$ & 90.0$^{+1.3}$ & 97.5$^{+1.4}$  \\
     
      & \textbf{$\tau$JP reg + DV-BASI$^*$}     & \textbf{87.5$^{+4.1}$} & \textbf{100.8$^{+5.4}$} & \textbf{89.1$^{+3.7}$} & \textbf{99.8$^{+3.7}$} & \textbf{92.3$^{+3.6}$} & \textbf{100.6$^{+4.5}$} \\
      \bottomrule
    \end{tabular}%
  }
  \label{tab:addition}
\end{table*}

Due to the iterative nature of DV-BASI, our method introduces some additional training time and storage overhead. While it is difficult to fairly compare all methods because of reproducibility issues, we emphasize that DV-BASI achieves comparable computational efficiency to current advanced approaches.
In terms of training time, compared with fine-tuning-based methods such as Adamerging \cite{AdaMerging} and aTLAS \cite{aTLAS} that both require learning sophisticated merging coefficients during training, our method directly takes the merged model as input and only introduces training time for refining it. Therefore, DV-BASI has comparable runtime to these methods.
For pre-training-based methods (e.g., $\tau$Jp Reg \cite{tau-jp}, Linearized Task Vector \cite{Tangent_Space}) that require substantial time to train each task-specific vector, DV-BASI is more efficient in terms of total time consumed.
Regarding storage, DV-BASI only needs to maintain one merged model and perform 3–5 refinement iterations. This results in storing only 4–6 models in total, which is significantly more resource-efficient than existing methods.
In summary, although DV-BASI introduces moderate additional computation, it still maintains a comparable and controllable level of computational cost.

\section{Experiments}

\label{others}

This section shows the effectiveness of improving task arithmetic performance by applying our DV-BASI algorithm under both supervised and unsupervised setting. 
Our experiments are conducted under both supervised and unsupervised conditions, \textbf{further experimental results and details are included in the Appendix}. For a supervised condition, we use normal cross entropy as our loss function. For unsupervised situations, follow Yang et al. \cite{AdaMerging}, we use entropy minimization as an optimization surrogate objective function to find the best group of coefficients each iteration (details included in Appendix).

\begin{table}[ht]
  \centering
  \footnotesize      

  \caption{Performance of test-time adaptation}
  \label{tab:addition-performance}
  \resizebox{0.28\textwidth}{!}{    
    \begin{tabular}{lcccccc}
      \toprule
      Methods & \multicolumn{2}{c}{ViT-B/32} & \multicolumn{2}{c}{ViT-B/16} & \multicolumn{2}{c}{ViT-L/14} \\
      \midrule
      Zeroshot   & 60.40 &        & 65.05 &        & 72.88 &        \\
      aTLAS      & 65.16 &        & 69.60 &        & 75.30 &        \\
      \textbf{DV-BASI}    & \textbf{66.76} & & \textbf{70.84} & & \textbf{76.01} & \\
      \bottomrule
      \label{tab:tta}
    \end{tabular}
  }
\end{table}

\begin{table*}[ht]
  \footnotesize
\caption{Results of applying DV-BASI after fine-tuning on different 20 tasks. CLIP with the ViT-B/32 backbone is used. }
\label{tab:singlemodel}
\centering

\resizebox{0.75\textwidth}{!}{
\begin{tabular}{lccccccccccc}
\toprule
\textbf{Datasets} & Cars & DTD & RESISC45 & SUN & Food101 & ImageNet & Caltech256 & PascalVOC & Country221 & UCF101  \\
\midrule
Finetune     & 78.26 & 78.94 & 95.94 & 75.40 & 88.58 & 76.41 & 92.60 & 88.42 & 21.99 & 85.01  \\
DV-BASI  & 80.30 & 78.94 & 96.05 &75.65 & 89.70 & 78.40 & 92.73 & 89.48 & 23.70 & 85.20  \\
\bottomrule
\end{tabular}
}

\resizebox{0.75\textwidth}{!}{
\begin{tabular}{lccccccccccc}
\toprule
  CIFAR10 & SVHN & CIFAR100 & FGVCAircraft & Flowers & OxfordPet & CUB200 & UCF101 & Caltech101 & EuroSAT & \textbf{AVG} \\
\midrule
    98.05 & 97.38 & 89.09 & 40.70 & 90.08 & 92.15 & 73.56 & 85.01 & 94.41 & 98.89 & 82.04 \\
 98.35 & 97.42 & 89.09 & 42.90 & 91.95 &92.21 & 73.56 & 85.14 & 94.47 & 99.70 & 82.74 \\
\bottomrule
\end{tabular}
}
\end{table*}

\paragraph{Datasets}
Following the typical evaluation procedure \cite{task_arithmetic,tau-jp, Tangent_Space, AdaMerging, aTLAS}, we focus on the computer vision task and apply our DV-BASI to $8$ image classification tasks: Cars \cite{Cars}, DTD \cite{DTD}, EuroSAT \cite{EUROSAT}, GTSRB \cite{GTSRB}, MNIST \cite{MNIST}, RESISC45 \cite{RESISC45}, SUN397 \cite{SUN}, and SVHN \cite{SVHN}. 
For task negation, we add ImageNet as a control dataset.

\paragraph{Compared Methods}
We conduct our experiment under supervised and unsupervised conditions. For supervised conditions, we choose aTLAS as our baseline method.
For unsupervised task-arithmetic based approaches, we choose training-free method (original task arithmetic with standard task vectors \cite{task_arithmetic}
and linearized task vectors \cite{Tangent_Space} and Ties-Merging \cite{TIES-Merging}) as our baseline method. AdaMerging \cite{AdaMerging} and aTLAS \cite{aTLAS} (We use the entropy minimization mentioned before to modify an unsupervised version of aTLAS) are selected as our Train-based baseline method.

\paragraph{Implementation}
The experiments are performed on NVIDIA GeForce RTX 4090 GPUs. For the initial model state $\theta_0$, we use our reproduced result on different baseline methods. 
Both supervised and unsupervised conditions are performed by taking ViT-B/32, ViT-B/16, and ViT-L/14 architectures in CLIP \cite{CLIP} as backbones.
All models are optimized by applying the AdamW \cite{adamW} optimizer with a learning rate of 0.01. 
To identify local minima, we adopt a commonly used criterion based on the model's performance on the validation/test set. 
Within $60$ epochs, we use early stopping with a patience of $5$ epochs to judge the best models (local optima points in parameter space), and then use such best model to update the difference vector to search continuously.

\paragraph{Task Negation}

Task negation aims to reduce the model's performance on a target task while maintaining performance on a control task. 
Following the standard evaluation procedure, the model is expected to \textit{forget as much as possible on the target task under the constraint of maintaining at least 95\% performance on the control task}. 

Denote the validation set for the target task by $\mathcal{D}_t$ and the control task by $\mathcal{D}_c$. We apply a simultaneous gradient descent on the control task and gradient ascent on the target task, thus the optimization problem in $(j+1)$-th iteration can be described as:
\begin{equation}
\begin{split}
\arg \min_{\Lambda_j} \; & \mathbb{E}_{(x,y) \in \mathcal{D}_t} \left[ -\mathcal{L} \left( f(x; \theta_j + \Lambda_j \delta_j), y \right) \right] \\
&+ \mathbb{E}_{(x,y) \in \mathcal{D}_c} \left[ \mathcal{L} \left( f(x; \theta_j + \Lambda_j \delta_j), y \right) \right].
\end{split}
\end{equation}

The findings presented in Table~\ref{tab:negation} demonstrate that DV-BASI significantly mitigates undesired biases (i.e., reducing performances of the target tasks) while sustaining over 95\% accuracy on control tasks. 

Under supervised and unsupervised conditions, for both linearized task vectors and standard task vectors, DV-BASI can effectively improve task negation and achieve SOTA performances.

\paragraph{Task Addition}
Task addition aims to create a multi-task model on several target datasets by adding task vectors from those target datasets. This operation allows us to transfer or reuse the knowledge from models individually fine-tuned on specific tasks.
Moreover, we embed the MOO algorithm we proposed and the regularization of $\tau$ JP into our DV-BASI in the cases of standard task vector and linearized task vector. 

As illustrated in Table~\ref{tab:addition}, DV-BASI enhances the performance of multi-task models merged through task addition. 
Specifically, DV-BASI achieves SOTA performance in both supervised and unsupervised task addition settings with both linearized and standard task vectors. 
Notably, when DV-BASI is applied to linearized task vectors, the relative accuracy may even exceed 100 percent. This highlights that DV-BASI fully utilizes the potential of task vectors and may even surpass the performances of fine-tuned models.

\paragraph{Test-Time Adaptation}
Test-time adaptation (TTA) \cite{liang2020we,sun2020test,wang2020tent} operates under the premise of absent labeled data for the target task, focusing on bolstering model robustness against domain shifts and out-of-distribution scenarios. Specifically, let \( T = \{ \tau^{(i)} \}_{i=1}^{n} \) represent the collection of task vectors for all accessible target sets, with \( \mathcal{D}^{(i)} \) denoting the corresponding dataset for task vector \( \tau^{(i)} \). For each target dataset \( \mathcal{D}^{(i)} \), the subset \( T \setminus \{ \tau^{(i)} \} \) is utilized to learn composition and mitigate knowledge leakage. We adhere to the experimental setup of \cite{aTLAS}, conducting offline adaptation on 22 image classification datasets (further details in the Appendix) with ViT-B/32, ViT-B/16, and ViT-L/14 architectures from CLIP \cite{CLIP} serve as our backbone models.
As shown in Table~\ref{tab:tta}, DV-BASI delivers superior performance across all models, achieving 66.76, 70.84, and 76.01, respectively, outperforming aTLAS by 1.60, 1.24, and 0.71 points. These findings underscore DV-BASI's proficiency in knowledge transfer, facilitating more effective adaptation in the absence of labeled data.

\paragraph{Enhancing Single-Task Performance Beyond Fine-Tuning} 

Inspired by the observation that the performance of the multi-task model merged by DV-BASI may outperform fine-tuned models, we wonder if DV-BASI is applicable to further improve model performance after fine-tuning on different datasets. Thus, we treat the fine-tuned weights of each dataset as the initial weights $\theta_0$ and apply DV-BASI to explore the possibility of further improvement. In our experiment, we chose 20 datasets from 22 datasets (due to the fine-tune accuracy on MNIST and GTSRB is already higher than 99 percent) in the previous TTA experiment and choose ViT-B-32 in CLIP \cite{CLIP} as backbone. All fine-tuned weights of 20 datasets are obtained from aTLAS \cite{aTLAS}. 

As shown in Table~\ref{tab:singlemodel}, the average accuracy among 20 datasets increases after applying DV-BASI. In detail, for 17 out of 20 tasks, model performances can be further improved. This demonstrates the potential of leveraging the difference vector for single task tuning.  
Therefore, we believe that edit and tune models with the difference vectors is very promising and has great potential for future development.

\section{Conclusion}
In this paper, we introduced DV-BASI, a novel multi-step optimization framework that extends the paradigm of task arithmetic by leveraging difference vectors as directed perturbations. Unlike traditional single-step arithmetic or parameter-efficient fine-tuning methods, DV-BASI offers a component-free and scalable way to continuously escape optimization stagnation through anisotropic scaling of difference vectors.
Our theoretical and empirical analysis highlighted several properties of DV-BASI. We conducted experiments across task arithmetic, test-time adaptation, and single-task tuning to demonstrate the validity of our approach with supervised and unsupervised objectives.


\section*{Acknowledgement}
The work was partially supported by the following: 
WKU 2025 Summer Student Partnering with Faculty Research Program under No. SSPF2025022,
WKU Internal (Faculty/Staff) Start-up Research Grant under No. ISRG2024009,
WKU 2025 International Collaborative Research Program under No. ICRPSP2025001.

\bibliography{aaai2026}

\end{document}